# Multi-View Graph Convolution Network for Internal Talent Recommendation Based on Enterprise Emails

**Abstract**

Internal talent recommendation is a critical strategy for organizational continuity, yet conventional approaches suffer from structural limitations, often overlooking qualified candidates by relying on the narrow perspective of a few managers. To address this challenge, we propose a novel framework that models two distinct dimensions of an employee's position fit from email data: WHAT they do (semantic similarity of tasks) and HOW they work (structural characteristics of their interactions and collaborations). These dimensions are represented as independent graphs and adaptively fused using a Dual Graph Convolutional Network (GCN) with a gating mechanism. Experiments show that our proposed gating-based fusion model significantly outperforms other fusion strategies and a heuristic baseline, achieving a top performance of 40.9% on Hit@100. Importantly, it is worth noting that the model demonstrates high interpretability by learning distinct, context-aware fusion strategies for different job families. For example, it learned to prioritize relational (HOW) data for 'sales and marketing' job families while applying a balanced approach for 'research' job families. This research offers a quantitative and comprehensive framework for internal talent discovery, minimizing the risk of candidate omission inherent in traditional methods. Its primary contribution lies in its ability to empirically determine the optimal fusion ratio between task alignment (WHAT) and collaborative patterns (HOW), which is required for employees to succeed in the new positions, thereby offering important practical implications.

**Keywords**
Graph Neural Networks (GNN); Multi-View Learning; Organizational Network Analysis (ONA); Talent Recommendation; Email Communication

# Chapter 1. INTRODUCTION

Timely talent acquisition is not just an operational task but a key driver of business success. When an employee suddenly leaves, the impact goes far beyond just an empty seat. The team loses the unwritten know-how and the informal connections that are vital for getting work done. This disruption immediately leads to serious problems, most notably a drop in productivity (Becker & Huselid, 1998). For this reason, the ability to quickly recruit a suitable candidate to fill such a vacancy is of critical importance to the business.

To fill roles quickly, companies generally rely on two main pathways: external hiring and internal mobility. While external hiring can bring in new skills and perspectives to an organization, internal mobility presents clear practical advantages, especially regarding candidate quality and the prompt onboarding for urgent positions. For example, a key study by Bidwell (2011) found that internal hires not only perform better at a lower cost than their external counterparts, but they also get up to speed faster and are more committed to the company. The strategic value of internal mobility is so recognized that global standards like ISO 30414 require companies to report the "internal hire rate" as a core metric.

Successful internal hiring depends on an accurate assessment of position fit. We argue that this fit comprises two core dimensions. The first, professional fit (the WHAT), encompasses an individual's knowledge, experience, and skills. The second, relational competency (the HOW), defines how they communicate, collaborate, and influence others within the organization. In traditional HR terms, WHAT corresponds to concepts like job family and job description, while HOW is represented by factors such as job level, job role, and communication or collaboration skills (Cable & DeRue, 2002; Sekiguchi & Huber, 2011). Despite its importance, however, conventional internal hiring processes often fail to reliably identify the most suitable candidates (Benson et al., 2019), leading to their frequent omission.

The existing internal hiring practices have fundamental limitations that often result in the omission of high-potential candidates and decision-making without sufficient deliberation. One root cause is a systemic issue stemming from opaque procedures and conflicting departmental interests. Hiring frequently takes place through informal managerial networks rather than formal job postings (Doeringer & Piore, 1971), leaving many employees unaware of opportunities to highlight their capabilities. This restricts the candidate pool to those already visible within small circles or informally endorsed, thereby excluding many qualified employees from consideration. Furthermore, managers who possess deep knowledge of their team members may be uncooperative. To avoid losing key performers, they may actively block transfers—a behavior known as "talent hoarding" (Haegele, 2022). This practice prioritizes short-term departmental goals over the broader objective of optimizing talent allocation across the organization. Caught between this data scarcity and lack of cooperation, already resource-strapped recruiters (Fisher et al., 2021) often have little choice but to bypass a comprehensive search and rely on a narrow set of recommendations.

Therefore, there is a pressing need for a method that allows HR to review all potentially suitable internal candidates quickly and comprehensively. Ideally, this method must assess not only a candidate's professional history but also their collaborative patterns, influence, and the contextual demands of the job itself. The goal is for HR to simulate the deep observational insight of a manager by conducting an exhaustive company-wide internal search, even in the absence of the manager's direct input.

In addition to procedural root cause, another limit lies in the nature of Human Resource Information System (HRIS) data: the absence of relational information. Organizations are dynamic and adaptive systems shaped by evolving interpersonal interactions (Schneider & Somers, 2006), yet HRIS data records static attributes of personnel, such as titles, departments, and job grades. While external hiring increasingly incorporates relational data through techniques like Graph Neural Networks (GNNs) (Frazzetto et al., 2025), such an approach is not fully feasible for internal hiring, which relies more on HRIS data with the static attributes of personnel. This limitation leads to two critical information gaps from the perspective of position fit. First, because an HRIS is not designed to capture the dynamic relations between members, it cannot fully reveal how they work, such as informal information flows or collaborative structures. This means the relational aspects that social network analysis (SNA) studies emphasize as core to organizational performance are entirely missing from the data (Cross & Parker, 2004). Second, even information about what they do is often limited to job titles or a job description, which is categorized and static. This fails to reflect the specific content of ever-changing projects and tasks, thereby limiting the ability to assess a candidate's actual job expertise.

This is just a conventional inevitability: any system based on static classifications is bound to incur information loss when applied to fluid, context-dependent work. Today, however, advances in data-driven technologies make it feasible to capture these dynamic aspects of work—beyond the static boundaries of traditional HRIS. A new approach is therefore required to extract such signals from within actual work artifacts.

To address these problems, we propose a novel framework that inductively extracts and integrates signals of WHAT (work content) and HOW (collaborative style, influence, communication competence) by leveraging email data, one of the most universal and naturally occurring data sources within an organization.

To combine these two perspectives, we apply multiple late fusion techniques, including simple concatenation, attention-based weighting, and a gating mechanism. These strategies allow the model to integrate both views into a unified representation that mimics a manager's holistic understanding of team members—without requiring direct human input.

Moving beyond static attributes like job titles or organizational charts, we formulate the task as a similarity-based node retrieval problem. Given a departed employee, the model identifies internal candidates in the latent space who are most similar based on both work content and collaboration behavior. To optimize this retrieval task, we apply pairwise ranking loss, enabling the model to rank candidates without requiring detailed job mapping or pre-specified vacancy profiles.

This approach enables us to ask and empirically test two core questions:

- RQ1: To what extent do the representations learned from the WHAT (semantic) and HOW (structural) perspectives align with formal HRIS data, thereby validating their effectiveness? Concurrently, how does the synergy between these two perspectives offer a more nuanced and dynamic understanding of organizational position than is possible with static HRIS data alone?

- RQ2: When representations from the two perspectives are integrated using various fusion strategies, how does the node retrieval performance (Hit@K) improve compared to single-perspective models, and which fusion strategy proves to be the most effective?

Through this process, our study makes several key contributions. First, it introduces a conceptual shift by redefining the notion of position fit: rather than relying on fixed labels, we model the WHAT and HOW dimensions as dynamic representations inductively derived from actual organizational behavior (Krasulja et al., 2016). Second, to address the challenge of label scarcity—a common issue in internal hiring settings—we leverage HRIS metadata such as job title, role, and level as proxy labels for weakly supervised learning. Finally, we enhance the explainability of our model through a gating mechanism that assigns context-specific weights to each type of information. This not only improves model adaptability but also yields interpretable outputs, aligning with the growing demand for explainable AI in HR applications (Ribeiro et al., 2016).

The remainder of this thesis is structured as follows. Chapter 2 reviews the theoretical background, including graph theory, network deep learning, and prior HR recommendation systems. Chapter 3 defines the research problem in detail. Chapter 4 presents the proposed model and experimental design. Chapter 5 reports and analyzes the results. Chapter 6 concludes with implications, limitations, and future directions.

# Chapter 2. LITERATURE REVIEW

**2.1. Graph theory and Graph Convolution Networks (GCN)**

A graph is a non-Euclidean data structure composed of vertices and edges, widely used to mathematically represent complex systems centered on relational properties. In this study, organizational email data are modeled as a structural graph capturing interactions between employees and a semantic graph representing the similarity of their work content. Unlike traditional tabular HR data, these graph-based approaches can reveal the dynamic context of on-the-job interactions, information that is typically difficult to capture.

Graph Convolution Networks (GCNs), as a type of Graph Neural Network (GNN), are deep learning architectures designed to learn the relationships defined over such graph structures, producing embeddings that jointly incorporate local structural patterns and node attributes. The GCN introduced by Kipf and Welling (2017) demonstrated strong representational learning capabilities in both supervised and unsupervised tasks by propagating structural context through multi-layer aggregation of first-order neighborhood information based on normalized Laplacian matrices.

A GCN is a powerful architecture that learns representations by integrating features not only from an individual employee but also from their connected neighbors. When applied to organizational data, this allows the model to directly utilize the principle of Homophily—the tendency for similar individuals to associate with one another. This process mirrors the '70:20:10' learning model (Lombardo & Eichinger, 2010), where a GCN's ability to learn from relational data effectively quantifies the informal learning that occurs through on-the-job experience and social interaction. Consequently, GCNs are particularly useful for internal talent recommendation as they can identify 'ready-made' candidates who have implicitly learned and performed similar roles through their work context, moving beyond merely matching explicit attributes.

In particular, when dealing with organizational data, GCNs are particularly valuable because they learn representations by incorporating not only the isolated attributes of individual employees but also those of their neighboring employees. This is especially important in the internal talent recommendation problem, as it allows us to identify individuals who are indirectly embedded in similar positions within the actual work structure, rather than simply sharing similar explicit attributes.

However, early GCN models faced key challenges, particularly with scalability on massive graphs and their uniform treatment of all neighbors. To tackle the scalability issue, researchers developed new sampling-based methods. GraphSAGE, for instance, samples a fixed number of neighbors for aggregation instead of using the full adjacency matrix (Hamilton et al., 2018), while FastGCN further optimized this for efficient mini-batch training and extended applicability to large-scale graphs through sampling-based neighborhood aggregation instead of relying on fixed adjacency matrices (Chen et al., 2018). Separately, to address the uniform treatment problem, the Graph Attention Network (GAT) introduced a mechanism to learn attention weights, allowing the model to assign different levels of importance to different neighbors during message passing (Veličković et al., 2018).

Through such diverse approaches aimed at enhancing the expressive power and scalability of structural learning, GCN-based models have continued to evolve. Building on these foundational principles, this study designs and compares several GCN-based models to find the right solution for the task.

**2.2. Organizational Network Analysis (ONA)-related studies**

Organizational Network Analysis (ONA) has long examined relational structures within organizations, drawing on core concepts from social network analysis and organizational sociology. Historically, however, ONA primarily relied on surveys and subjective assessments, which constrained its scalability and limited its ability to capture dynamic processes over time. The advent of Human Resource Information Systems (HRIS) and the proliferation of digital communication records have developed this field, enabling large-scale, data-driven approaches to ONA that leverage graph-based techniques to address organizational challenges far beyond the scope of static analysis.

Early studies sought to use static network properties to predict individual outcomes. Cross & Parker (2004), for instance, demonstrated in their practical work applying ONA to real organizational problems that the quality and pattern of relationships—such as how quickly one can access necessary information (closeness) and connect diverse groups without becoming a bottleneck (brokerage)—are correlated with performance. Yuan et al. (2015) empirically demonstrated that relational indicators in email networks—such as in-degree and clustering coefficient—are significantly correlated with employees' turnover and promotion, illustrating that network structures themselves, beyond individual attributes, can serve as meaningful predictors of career mobility. Building on this, Gloor et al. (2017) observed that managers often experience declines in centrality before resigning, suggesting that temporal shifts in centrality can function as early signals of leadership gaps. As a result, centrality-based analysis has become foundational for understanding and forecasting organizational dynamics.

Centrality measures have also been widely used to identify leadership and influence. Knaub et al. (2018) examined peer-nominated leaders within an educational organization and found that most leaders emerged as hub nodes with high direct of indirect connectivity, empirically supporting theories that social influence arises from structural connectedness. Extending this perspective, Haemers (2021) analyzed corporate email networks and revealed that individuals not formally in senior positions could still occupy highly central roles—highlighting both the limitations of relying solely on formal hierarchies and the power of ONA to quantify informal competencies that often escape traditional HR evaluations.

Recent work has increasingly focused on uncovering "embedded roles" that are not fully explained by structural connectivity alone. Barnes et al. (2024), for example, analyzed long-term email logs to quantify how organizational hierarchies shape communication, finding that higher-ranking members not only received but also sent more messages, resulting in an overall upward flow. While this partially aligns formal rank with communication power, consistent with prior research, it also reveals embedded roles and underscores that explicit hierarchies do not fully capture actual communication dynamics.

Taken together, these studies indicate that while traditional centrality measures remain powerful for identifying talent and leadership roles aligned with formal structures, learning-based approaches that reveal hidden communication patterns and emergent roles beyond conventional frameworks offer clear advantages. From this perspective, our study introduces a recommendation framework that richly combines centrality metrics, the structural network itself, and semantic attributes and networks to identify both roles rooted in existing hierarchies and those implicitly embedded in the network that traditional views may miss.

**2.3. Semantic embedding approaches for talent exploration**
Traditional talent recommendation systems have largely relied on rule-based matching using structured, resume-derived attributes. However, such approaches often struggle to capture the diversity and fluidity of actual work practices in organizational settings. For instance, individuals holding the same job title may perform substantially different tasks or exhibit distinct collaboration patterns—subtleties that structured data stored in Human HRIS alone typically fail to discern.

In this context, recent studies have turned to analyzing unstructured textual data such as emails and resumes to quantify the content and contextual nuances of work. Zhang et al. (2021), for example, introduced EmailSum, a model that automatically summarized lengthy email threads and extracts sender intent and roles (e.g., decision maker, coordinator). This work closely parallels our approach of leveraging email embeddings to construct semantic networks that reflect an individual's work context.

There is also a growing body of research on matching talent by embedding concepts such as job roles, positions, and skills alongside unstructured data found in resumes and job postings. Zhao et al. (2021) embedded large-scale datasets of resumes and job descriptions to build a dense retrieval system that performs a two-stage matching, focusing on contextual similarity to simultaneously improve recommendation accuracy and click-through rates. Similarly, Lavi et al. (2021) employed SBERT to match resumes with job postings, demonstrating more precise semantic alignment than traditional keyword-based TF-IDF approaches. Bocharova et al. (2023) further advanced this line by projecting roles and skills into a shared embedding space, enhancing matching by leveraging mutual semantic similarity through their framework.

Semantic embeddings thus serve as a powerful tool for quantifying similarity at the level of work context, far beyond simple keyword overlaps. While existing studies have primarily addressed isolated problem formulations, our work moves beyond mere pairwise semantic similarity comparisons. We construct entirely new edges—semantic similarity-based links—between nodes, enabling information to flow along these connections and integrating the resulting network with a structural graph into a unified learning framework. In doing so, we offer a distinctive contribution by enriching relational representations and revealing talent connections that traditional approaches would overlook.

**2.4. GNN approaches for Human Resources recommendation**

Complex HR datasets—comprising diverse entities and intricate interactions both within and beyond organizational boundaries—are inherently well-suited to graph-based modelling, which captures relational structures in a learnable form. This section therefore reviews not only simpler graph and semantic embedding studies introduced earlier but also various GNN approaches in the HR domain, along with systems that could evolve into such models. We also situate our work within this technical landscape by classifying it according to graph construction and GNN architecture.

Notably, De Vos et al. (2024) developed an internal mobility recommender system that modeled employee-to-employee performance similarity through a collaborative filtering framework implemented on a graph structure. By incorporating similar regularization, they further aligned latent representations based on personal attributes. Although the work did not employ GCNs or similar architectures, it stands as a good example of leveraging graph-based relationships for HR recommenders, especially within internal mobility context.

More explicit applications of GNNs typically begin by directly learning from structural signals such as collaboration frequencies or interaction intensities. Hu et al. (2022), for instance, modeled team restructuring as a cluster graph with edges reflecting intra-organizational collaboration and message exchanges, applying GCNs to identify alternative team formations. Yang and Shen (2025) used a single GCN framework to predict an individual's future capabilities, structuring talent-job-skill relationships into a knowledge graph to form personalized competency maps. Our work also starts by independently training GCNs but diverges by later merging these structural representations with semantic aspects through a dedicated integration phase.

Beyond capturing structural topologies, several studies have embedded semantic information—often derived from resumes or job descriptions—into graph learning. Wasi (2024), for instance, extracted entities from textual data using a large language model, embedding them as BERT feature nodes within a knowledge graph, which a GNN then used for job categorization and centrality-driven recommendations. Xue et al. (2025) conducted binary classification for resume-job matching by applying GNNs over resume graphs equipped with BERT embeddings, while Li et al. (2025) incorporated GPT-4o embeddings. Our work similarly uses word2vec embeddings as node features but moves beyond by introducing explicit semantic similarity thresholds to form new edges, thereby converting latent contextual closeness into concrete pathways that propagate homogeneity within the graph.

Research has also actively explored integrating multiple information sources using graphs. Many studies employ heterogeneous graphs that mix edge or node types directly inside the GNN, whereas some adopt explicit fusion outside the network. Li et al. (2025) passed GPT-4o embeddings though a hierarchical GNN, sequentially processing layers of resumes, jobs, and companies. Hou et al. (2022) embedded candidate-attribute-job triples and textual semantics into a heterogeneous GNN, learning separate attentions over distinct relation types—such as candidate-job or candidate-skill—to maintain interpretability of the interactions. Behar et al. (2024) similarly distinguished edge types like applications, evaluator feedback, and surveys, applying tailored attentions to capture their varying contributions. Hue et al. (2025) diverged slightly by fusing local semantic features (BERT-encoded resumes) with global structural features (GCN-learned similarity graphs) only after separate encodings, using a knowledge-aware attention network (KAN) to dynamically weight these complementary signals.

In contrast, our method fully separates the learning of structural and semantic graphs, subsequently combining them through adaptive gating that modulates the relative importance of each dimension at the level of individual node pairs. By constructing a new semantic similarity network and training it in parallel with a frequency-based structural graph, our approach allows the model to learn how different facets of organizational signals should be weighted, achieving a finer granularity than implicit fusions common within GNNs.

Finally, LinkSAGE (Liu et al., 2024) stands as a benchmark for industrial-scale GNN deployment in HR contexts. Operating over a massive heterogeneous graph of users, jobs, skills, and companies, it unified these diverse entities via message passing, learned from explicit user interactions (applications, views), and coupled GraphSAGE encoders with Deep Neural Network (DNN) decoders in a nearline architecture to balance low latency and large-scale operability.

Unlike these large engineering systems, our work centers on architectural contributions tailored to the unique challenges of HR data. By exploiting weak supervisory signals embedded in organizationally validated roles and positions—particularly in internal talent identification scenarios lacking explicit behavioral labels—our approach can mine raw email logs to identify viable candidates, offering a practical alternative that minimizes engineering overhead while enabling straightforward experimentation and deployment.

# Chapter 3. METHODOLOGY

**3.1. Data description and preprocessing**

3.1.1. Dataset overview

This study uses email log data from a mid-sized company. The dataset covers approximately six months and includes 192,537 email exchanges involving 1,518 employees. To ensure confidentiality, all personally identifiable information was removed: employee IDs were replaced with randomly generated codes, and only the subject lines—excluding full email bodies—were retained for analysis.

Table 1. Structure of the email data

| Structure | Description |
| --- | --- |
| Node | Unique identifier assigned to each employee involved in sending or receiving emails |
| Edge | Historical records of email transmissions between pairs of nodes (weighted by frequency) |
| Attribute | Subject lines of emails exchanged between nodes |

Figure 1 below illustrates how email data inherently contains both structural and semantic dimensions. Beyond simply reflecting the network of message transmissions—depicted by blue edges— it also embeds the substantive nature of work through the content of subject lines, shown as orange-envelop icons. This dual characteristic provides a foundation for modeling both interactions within the organization [1]. Accordingly, this study refines and exploits these two categories of signals by representing them as graph views and feature sets.

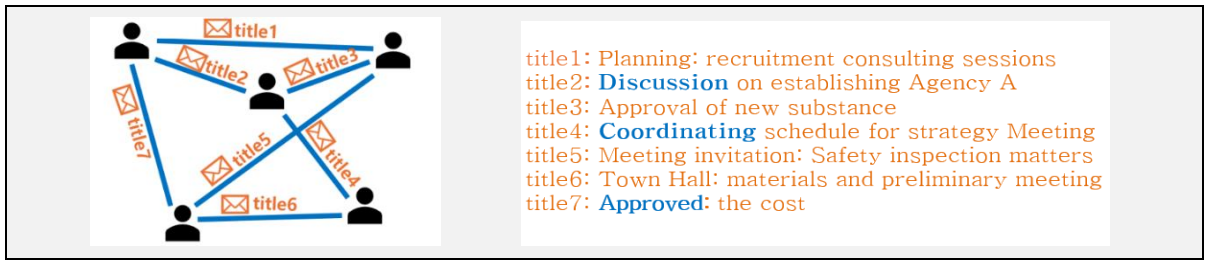

Figure 1. Email data's structural and semantic information

Notably, node attributes were intentionally excluded from use as direct learning features. This design reflects a methodological choice to explore employee similarities in a bottom-up manner, independent of any pre-defined organizational taxonomies. Traditional HR attributes such as job families and roles were instead employed for exploratory analyses, including comparisons with dynamic constructs like clusters derived from embeddings and network structures. Details of these analyses are provided in section 5.1.

3.1.2. Embedding of email subject lines

Given the nature of this study—extracting job and role characteristics from short subject lines typically comprising 5 to 10 words— we prioritized capturing the intrinsic meaning of individual terms over broader contextual nuances. Consequently, we opted for word-level embeddings using Word2Vec, rather than sentence-level models such as BERT or SBERT. Additionally, to enrich the diversity of domain-specific language and enhance the generalizability of the embeddings, nouns extracted from internally maintained job descriptions were incorporated into the training corpus.

---

[1] Certain text expressions in email subjects (e.g., "coordinate," "discuss", "approve") may also contain structural cues such as the node's role or relational centrality; however, these are considered minor signals. Rather than explicitly distinguishing these subtle cues, the model is designed to allow them to be learned naturally through feature interactions during the subsequent graph embedding process.

Each subject line was processed by extracting nouns and tokenizing on whitespace, then transformed into fixed-length 100-dimensional vectors via a Word2Vec model. These vectors were subsequently aggregated via mean pooling into a single dense vector representation of dimension 100. The model was trained using Gensim's skip-gram implementation.

Table 2. Examples of embedding process

| Email subject | ['payroll', 'adjustment', 'error', 'confirmation', 'request'] |
|---|---|
| Embedding process | {'payroll': [0.1624, -0.3841, … , 0.5189, -0.2496], 'adjustment': [-0.3012, 0.0973, … , 0.1487, 0.4172], 'error': [0.5650, -0.1229, … , -0.3794, -0.2186], 'confirmation': [-0.0876, 0.3321, … , 0.2079, -0.1443], 'request': [0.2210, -0.2988, … , -0.0604, 0.0768]} |
| Mean-pooled result | {'payroll adjustment error confirmation request': [0.11192, -0.07528, …, 0.08714, -0.0237]} |

3.1.3. Filtering and preliminary text preprocessing

Preprocessing was carefully aligned with this study's objectives. Emails were excluded if they mentioned company names, specific individuals, personal data, or were system-generated without a human sender or recipient. For subject lines mixing Korean and English, Korean nouns were extracted as above, and English words tokenized by whitespace. To refine the corpus, TF-IDF was used to filter out low-information terms, and words in the top and bottom 5% of frequency were removed. Expressions indicating dates or times, uniformly distributed across roles and teams, were also excluded to prevent confounding.

**3.2. Graph construction**

This study builds two distinct graphs from the email logs, each capturing a different aspect of work relationships. These graphs are later combined within a multi-view framework as inputs to the GNN models.

3.2.1. Structure Network

The Structure Network connects employees who have exchanged emails, with edge weights reflecting the total frequency of their interactions. It ignores the content of communications, focusing solely on the existence and strength of connections.

Table 3. the Structure Network

| Structure | Description |
|---|---|
| Node | Unique identifier assigned to each employee |
| Edge | Connects pairs of nodes $\{V_i, V_j\}$ based on the existence of actual email exchanges |
| Graph Type | Undirected, weighted graph |
| Definition | $G_{str}(V_i, V_j) = Count(V_i, V_j) + Count(V_j, V_i)$ |

3.2.2. Semantic Similarity Network

The Semantic Similarity Network captures task-related similarities by modeling semantic relationships derived from email subject lines. Unlike typical GNN-based social recommendation systems that create explicit node-item interactions, it infers indirect ties by projecting edge-level semantic similarities onto node pairs. A cosine similarity threshold of 0.75 (near the median of the embedding distribution) was empirically chosen to ensure meaningful similarity levels and stable model performance. Raw similarity scores were directly used as edge weights and scaled to 0.5–1.0 to improve training stability.

Table 4. the Semantic Similarity Network

| Structure | Description |
|---|---|
| Node | Unique identifier assigned to each employee |
| Edge | Edge weight is the cosine similarity between the representative semantic embeddings $(s_i, s_j)$ of two nodes. An edge is formed only if the similarity is above a threshold τ. |
| Graph Type | Undirected, weighted graph |
| Definition | $G_{ssim}(V_i, V_j) = Sim(s_i, s_j) \cdot I(Sim(s_i, s_j) \geq \tau)$ |

As detailed in Section 3.1.2, each email subject was encoded as a 100-dimensional embedding. To reframe this as a node-level similarity problem, we obtained node-level representations by simply computing the average of all email embeddings associated with each node, yielding a semantic centroid for each employee. Let $s_i$ denote the representative embedding for employee V_i, and Sim(α,β) the cosine similarity between vector α and β.

3.2.3. Multi-view graph architecture

The two graphs capture fundamentally different aspects of work relationships. In the Structure Network, edges exist only where direct communication occurred, with edge strength reflecting interaction frequency. In contrast, the Semantic Similarity Network connects nodes with strong edges even without direct exchanges, as long as their tasks(emails) are semantically similar. The recommendation model integrates these complementary perspectives within a unified multi-view GNN framework.

**3.3. Node features**
In this study, each node is represented by a combination of structural and semantic information, serving as a rich input to the recommendation algorithm. The features used are detailed below.

3.3.1. Semantic embedding

The semantic embedding for each node was calculated as the average of the embeddings of the email subjects linked to that node, effectively treating these subject lines as indicative of its task profile. Cosine similarity was then used to measure the closeness between these embeddings, forming a central component in generating recommendation scores. This vector corresponds to the representative embedding $s_i$, introduced earlier in Section 4.2.2.

3.3.2. *Centrality*

Each node also carries structural attributes that reflect its position and role within the communication network. These were incorporated as structural features in the recommendation algorithm. To quantitatively capture how they interact, this study employed four key centrality measures.

*First, degree centrality* counts the number of edges directly connected to a node, calculated as:

$$d_v = deg(v)$$

This captures how actively an individual participates in organizational exchanges, reflecting both communication frequency and the extent of their contact network. It is particularly helpful for identifying hubs that handle substantial volumes of information.

*Second, closeness centrality* is defined as the inverse of the sum of the shortest distances to all other nodes, fcalculated as:

$$c_v = \frac{1}{\sum_{\{u \neq v\}} d(v,u)}$$

where $d(v,u)$ denotes the shortest-path distance between node $v$ and $u$. It indicates how near a node is to the center of the network. A higher closeness value suggests the node can quickly reach others, making it a useful indicator for identifying employees with broad communication reach or the potential for rapid information spread.

*Third, betweenness centrality* measures how often a node lies on the shortest paths between other nodes. Its formula is:

$$b_v = \sum_{\{s \neq v \neq t\}} \frac{\sigma_{st}(v)}{\sigma_{st}}$$

where $\sigma_{st}$ is the total number of shortest paths between nodes $s$ and $t$, and $\sigma_{st}(v)$ is the number of those paths that pass-through node $v$. It quantifies a node's role as an intermediary in the network's information flow. Nodes with high betweenness are likely to act as brokers, bridging different groups and facilitating cross-boundary communication.

*Lastly, eigenvector centrality* assesses a node's importance not only by its number of connections but also by the influence of its neighbors. It is defined by the relation:

$$e_v = \frac{1}{\lambda} \sum_{\{u \in N(v)\}} w_{vu} e(u)$$

where the centrality of node $v(e_v)$ is proportional to the sum of the centralities of its neighbors ($e_u$), and $\lambda$ is the largest eigenvalue of the graph's adjacency matrix. It operates on the principle that connecting to well-connected nodes boosts a node's own significance, offering insights into its overall standing and influence. Nodes with higher eigenvector scores are typically linked to other key players, exerting broad structural influence.

### 3.3.3. Concatenated features

Although the semantic embedding vectors from Section 4.3.1 and the centrality metrics from Section 4.3.2 originate from different data structures, in this study, they are concatenated into a single node feature vector and used as a unified input. These node features are then learned within a GNN framework through the edge structures of the two previously defined graphs (the Structure Network and the Semantic Similarity Network) and are updated simultaneously through the message-passing process.

By combining these two types of features into a single node representation, it becomes possible to complementarily describe both WHAT an employee does (their work content) and HOW they do it (their work style). This enables a search for position fit candidates who not only perform tasks similar to those of the departed employee but also possess a similar role and communication structure within the organization.

### 3.3.4. Exploratory analysis of node features

The bottom-up approach proposed in this study is data-driven, unlike existing static filtering approaches. However, to establish the reliability and validity of the proposed model for real-world application, it is essential to verify its consistency with the existing HR classification system, namely job family and role. If the proposed features demonstrate a certain level of alignment with the traditional classification system, it suggests the model's potential as a more sophisticated recommendation tool that integrates relational information.

For this analysis, we used the Job Family and Role columns from existing HR data. (These are also later used as weak labels.) The analysis consists of three parts. First, the silhouette score and AUC were calculated to assess how well each feature can distinguish between job families and roles. Second, the performance of various combinations of the five features in predicting job families and roles was evaluated; through this, we compared the performance difference between using only semantic embeddings and only centrality metrics to examine the complementarity of the two types of information. Finally, the alignment of the information-rich semantic embeddings with the existing job family classifications was intensively analyzed.

### 3.4. Score-based heuristic model (baseline)

#### 3.4.2. Baseline configuration

In this study, a heuristic, score-based baseline model was designed for performance comparison against the deep learning-based recommendation model. According to results from prior feature combination experiments, semantic embedding alone demonstrated high classification performance, while centrality metrics provided a certain level of performance improvement in a supplementary role.

Accordingly, semantic similarity, representing job and task content, was established as the primary criterion, with the differences between centrality metrics applied as a complementary difference term (or penalty term). The final recommendation score is calculated as follows:

$$Score(V_i, V_j) = \alpha_s \cdot Sim(s_i, s_j) - \alpha_d \cdot |d_i - d_j| - \alpha_c \cdot |c_i - c_j| - \alpha_b \cdot |b_i - b_j| - \alpha_e \cdot |e_i - e_j|$$

$Sim(s_i, s_j)$ is the cosine similarity between the mean Word2Vec-based embeddings, while $d_i, c_i, b_i, e_i$ represent the degree, closeness, betweenness, and eigenvector centrality for node $V_i$, respectively.

To reflect the primary importance of semantic information, the weights were determined through experimentation and set as follows:

$\alpha_s$ = 0.8, with the weights for each centrality penalty term set equally to $\alpha_d = \alpha_c = \alpha_b = \alpha_e$ = 0.05.

This model is designed to center on semantic similarity while penalizing candidates who are structurally too dissimilar, thereby reflecting a balance between actual task similarity and organizational activity patterns.

#### 3.4.3. Advantages and limitations of the score-based ranking approach

The score-based recommendation approach has the advantages of simple implementation and high explainability. Because each recommendation score provides a clear numerical basis for why a candidate was recommended, it allows for intuitive interpretation by decision-makers when used in actual HR processes.

However, limitations also exist. Since centrality features are condensed into single scalar values, they cannot reflect multidimensional relationships within the network structure (e.g., community structures). Furthermore, non-linear interactions among the five features are not captured, making it difficult to fully account for complex tasks and roles characteristics where semantics and structure are intertwined. These limitations stem from the fundamental structural constraints of an approach that treats individual features as fixed values and combines them through a weighted sum. The expressive limitations of this score-based method can be addressed by the GNN-based model presented in the next section, which learns semantic information and structural relationships between nodes in an integrated manner.

**3.5. GNN-based recommendation model**

3.5.1. Advantages of adopting a GNN Structure

A GNN can learn representations that integrate the complex context of both structural position and semantic information by iteratively learning not only the features of each node but also its connection relationships with adjacent nodes. Furthermore, it has the advantage of extensibility, allowing for the flexible integration of various types of relationships (edge types) and diverse node features, and can dynamically adjust the strength of influence between nodes through attention or gating mechanisms. In conclusion, a GNN is a highly viable structural modeling approach for this problem, as it can integrally learn an individual's task similarity and their communication patterns within the organization.

3.5.2. The artificial introduction of the Semantic Similarity Network

Unlike the Structure Network, which is predicated on physical connections based on collaboration records, the Semantic Similarity Network is generated based on the similarity between the mean vectors of email subject embeddings that each node has handled. It artificially creates connections between nodes with high semantic similarity, even if no actual interaction exists between them.

This configuration is closely related to the concept of Homophily (McPherson et al., 2001), a frequently discussed topic in traditional network science, which posits that connections are more likely to form between nodes with similar characteristics. The Semantic Similarity Network "artificially engineers" this homophily, thereby providing an inductive bias for the GNN to adjust the embedding representations of employees with similar job content to be closer to each other. (In contrast, the Structure Network is a network formed based on actual collaboration frequencies and paths, reflecting the practical information flow and structural context within the organization through network centrality metrics like hubness and brokerage roles.)

These two graphs, constructed in different manners, simultaneously provide different types of relationships (semantic vs. structural) and serve as the structural foundation for integrating complementary information during the GNN's representation learning process. For example, let's assume Employee A, who was in charge of domestic legal affairs, has left the company.

Table 5. Hypothetical case comparing candidates for position A

| Comparison item | Departed employee A | Candidate ① | Candidate ② |
|---|---|---|---|
| Job title | Domestic Legal | Domestic Legal | International Legal |
| Team name | Legal Team 1 | In-house Counsel Team | Compliance Team |
| Team characteristics | 4 Domestic, 5 Intl. Legal | Sole practitioner at a small subsidiary | 9 Domestic, 1 Intl. Legal |
| Main collaborators | Numerous domestic legal professionals | (Few collaborations due to solo work) | Numerous domestic legal professionals |

Based solely on Semantic Similarity, Candidate ①, who has handled many topics related to domestic legal affairs, might receive the highest similarity ranking. However, if Candidate ① is isolated within the Structure Network, the GNN cannot perform message passing using information from neighboring nodes. Consequently, it must learn the embedding for that node relying only on its own initial features (self-feature). On the other hand, if Candidate ② is structurally close to other domestic legal professionals, the GNN's message passing will reflect the features of these neighboring experts in Candidate ②'s embedding, gradually adjusting the two embedding vectors in a similar direction.

As a result, the model proposed in this study moves beyond looking only at explicit job titles and roles to capture latent suitability revealed through actual tasks and collaborative relationships. This aligns with the perspective of the '70:20:10' learning model (Lombardo & Eichinger, 2010) mentioned earlier, which suggests that 70% of competency development occurs through on-the-job experiences and interactions. Although Candidate ②'s official job title is different, it is highly probable that they have already been acquiring practical job competencies through close collaboration with domestic legal experts. Therefore, this model provides an objective basis for recommending Candidate ②, whose job title differs, as a suitable successor alongside the Candidate ①.

3.5.3. GNN input structure and fusion strategies

The input data and fusion strategy for the GNN model used in this study were designed as follows.

Node feature vector. The feature vector for each node (employee), used as input for all models, is constructed by combining semantic and structural features. Specifically, the 100-dimensional Word2vec-based email subject embedding is denoted as s_i, and the four centrality metrics—degree, closeness, betweenness, and eigenvector—are defined as $d_i, c_i, b_i, e_i$, respectively. The final input feature vector $x_i$ is generated by concatenating all of these components:

$$x_i = Concat(s_i, d_i, c_i, b_i, e_i) \in R^{104}$$

Graph structure. The model uses two types of networks as input: the Structure Network, based on actual interactions, and the Semantic Similarity Network, based on the semantic similarity of work content. These are utilized independently on this experimental setup.

Fusion strategy. To integrate information from the two graphs, six fusion strategies were designed, ranging from single-graph models to various fusion techniques. The performance of all methods was compared based on the same node features and loss function (pairwise ranking loss).

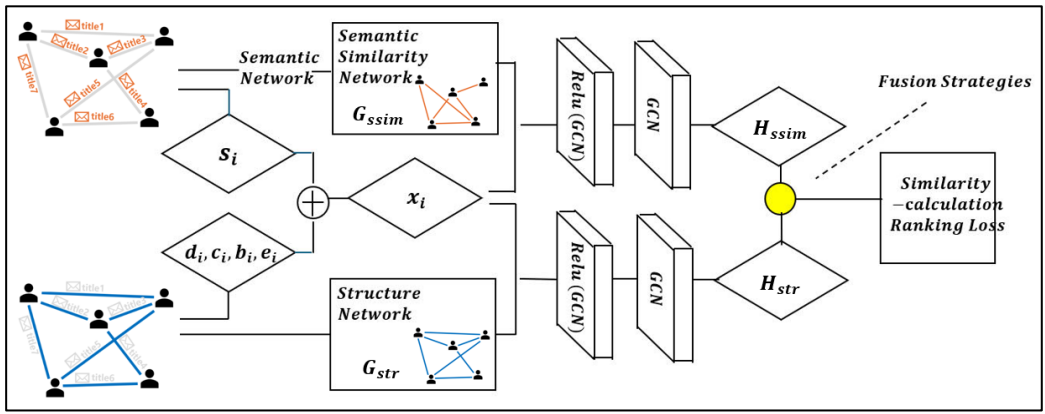

Figure 2. Model architecture example (late fusions)

Table 6. Comparison of fusion strategies

| Strategy | Description | Formulation |
|---|---|---|
| ① Single GCN | Each graph is learned independently. | $H_{str} = GCN(x, G_{str})$<br>$H_{ssim} = GCN(x, G_{ssim})$ |
| ② Early Fusion (Concat) | The adjacency matrices of the two graphs are first concatenated, then fed into a single GCN. | $G_{early} = (G_{str}, G_{ssim})$<br>$H_{early\_concat} = GCN(x, G_{early})$ |
| ③ Late Fusion (Concat) | Embeddings learned independently from each graph are concatenated channel-wise. | $H_{late\_concat} = Concat(H_{str}, H_{ssim})$ |
| ④ Late Fusion (Weighted sum) | The two embeddings are combined using a pre-defined hyperparameter α for a weighted sum (α= 0.8) | $H_{weighted} = \alpha H_{str} + (1-\alpha) H_{ssim}$ |
| ⑤ Late Fusion (Attention) | An attention score $\alpha_v$ reflecting the importance of the two embeddings is learned for each node to apply dynamic weights | $h_{i,att} = \alpha_i h_{i,str} + (1-\alpha_i) h_{i,ssim}$ |
| ⑥ Late Fusion (Gating) | A gate vector $g_i$ is learned to determine the mixing ratio of the two embeddings, enabling fine-grained control over information flow. (Here, σ is the sigmoid function $W_g$ and $b_g$ are learnable parameters, and ⊙ denotes the element-wise product.) | $g_i = \sigma(W_g[h_{i,str}\|h_{i,ssim}] + b_g)$<br>$h_{i,gated} = g_i \odot h_{i,str}$<br>$\quad\quad +(1-g_i) \odot h_{i,ssim}$ |

The score-based model from Section 3.4 has limitations in learning complex non-linear relationships due to its linear structure. To overcome this, a GNN was introduced. To validate its effectiveness and to explore the optimal fusion architecture for effectively combining the two information sources (semantic, structural), we designed a series of phased comparative experiments as outlined above. Furthermore, this study aims to empirically determine which fusion strategy most effectively integrates the two information sources for the talent recommendation problem.

First, the ① Single GCN model serves to verify whether a basic GNN structure applied to a single information source (either structural or semantic) can surpass the expressive limits of the score-based baseline model. Next, to explore the impact of combining the two different types of information sources, we evaluate the performance of ② Early Fusion (Concat) and ③ Late Fusion (Concat), which physically combine the information without additional learnable parameters. Taking a step further, the ④ Weighted Sum method explores the effect of the researcher directly controlling the importance of the two information sources with a hyperparameter.

However, these static or simple fusion methods have a fundamental limitation in that they apply the same rule to all nodes. To overcome this, we introduced adaptive mechanisms that allow the model to learn the optimal fusion method directly from the data. ⑤ Attention dynamically controls the information flow at the node-level, while the most sophisticated model, ⑥ Gating, does so at the feature-level. These adaptive models have the potential to learn the complex and contextual interactions that the score-based model could not capture.

### 3.6. Training and evaluation

This study defines the internal talent or succesor recommendation problem as an open-ended task and trains the model by leveraging weak supervision signals in a situation where explicit ground truth labels do not exist. The recommendation model is structured to suggest candidate B who can perform a similar role when a specific employee A departs from the organization. The training was conducted based on a pairwise ranking loss between pairs of nodes.

3.6.1. Construction of weak labels

A positive pair was defined as a pair of nodes sharing the same job family and the same role. When both conditions were met, the match was considered appropriate, with sufficient position similarity for that position being secured. This is based on the practical judgment that while internal talent recommendations are not always made according to formalized criteria, the alignment of job family and role can function as a reasonable proxy for assessing a certain level of position similarity in early talent-pooling processes. (Furthermore, the utility of this practical judgment is exploratorily analyzed in Section 5.1.)

This label construction does not represent absolute ground truth; it is designed to act as a weak supervisory signal necessary for the model to learn task similarity and organizational activity patterns. Conversely, a negative pair was constructed by randomly sampling from pairs of nodes that do not meet this condition.

A margin-based pairwise ranking loss was used for training, configured so that the model learns to assign a higher score to positive pairs than to negative pairs. This approach, which learns the relative suitability among candidates, naturally corresponds to the comparative decision-making structure of talent evaluation in real-world HR scenarios.

3.6.2. Quantitative evaluation metrics

Model performance evaluation was based on whether a positive candidate was included within the top-K recommended results. The primary metric used was Hit@K, with K set to 30 and 100. For each departing node in the test set, recommendation scores were calculated for the entire candidate pool, and the proportion of cases where a positive candidate was included in the top 30 and top 100 was measured. The values for K were chosen to reflect the scope of a candidate pool that is feasible for short-term review in practical HR settings and were used as a metric to evaluate ranking-based recommendation performance.

# Chapter 4. RESULTS

**4.1. Feature exploration and validation**

In this section, we validate the effectiveness of the node features designed in Chapter 4 and exploratorily analyze how the proposed bottom-up approach relates to the actual organizational structure.

4.1.1. Analysis of the discriminative power of individual features

Table 7. Clustering and separability analysis of individual node features

| feature | Silhouette Score (Job Family) | AUC (Role) |
|---|---|---|
| $s_i$ (semantic embedding) | 0.338 | 0.308 |
| $d_i$ (degree) | -0.124 | 0.697 |
| $c_i$ (closeness) | -0.120 | 0.713 |
| $b_i$ (betweenness) | -0.209 | 0.659 |
| $e_i$ (eigenvector) | -0.188 | 0.651 |

The analysis of individual features in Table 7 reveals that semantic information (WHAT) and structural information (HOW) capture distinct dimensions of an employee's organizational identity. The semantic embedding($s_i$) effectively discriminated between job families, achieving a Silhouette Score of 0.338. However, it showed limited discriminative power for Roles.

Conversely, all four centrality metrics proved effective at separating Roles, with AUC scores ranging from 0.651 to 0.713. Yet, they failed to distinguish Job Families (the WHAT dimension), yielding negative Silhouette Scores (from -0.120 to -0.209) that indicate virtually no discriminative power in this regard.

4.1.2. Predictive performance evaluation of node feature combinations

Table 8 presents the F1-scores for classification models using 31 distinct feature combinations, offering deep insights into how semantic (WHAT, $s_i$) and structural (HOW, centrality) information interact to predict the established categories of Job Family and Role.

The dominant predictive power of semantic information. The analysis first highlights the dominant influence of the semantic embedding ($s_i$) across both prediction tasks. The model using $s_i$ alone (#16) achieved a high F1-score of 0.8802 for Job Family prediction. Notably, it also yielded a strong F1-score of 0.7539 for Role prediction—significantly outperforming the model that combined all four structural features (#17, F1-score of 0.4892). This result contrasts with the low AUC (0.308) from the simple separability analysis in 5.1.1, proving that while roles are not linearly separable in the semantic space, the embeddings contain rich latent patterns that a classifier can effectively learn.

The synergistic effect of structural information . While semantic information provides a powerful foundation, the results also reveal the synergistic effect of structural information. Although the $s_i$-only model (#16) was strong, adding all structural features (#1) achieved the peak F1-score of 0.9012 for Job Family prediction. Similarly, for Role prediction, the highest performance (0.7693) was attained by specific combinations of semantic and structural features. This demonstrates that considering structural context— such as an individual's influence or brokerage role within the organization—provides complementary information that refines and enhances predictions, leading to a more complete understanding of an individual's position fit.

In conclusion, this analysis strongly suggests that semantic information provides a robust foundation for identifying candidates, while structural information acts as a complementary partner that enables more sophisticated, multi-dimensional judgments. This finding provides a core empirical justification for our GNN model's design, which aims to integratively learn from both information sources.

Table 8. Macro F1-scores by feature combinations

| No. | $s_i$ | $d_i$ | $c_i$ | $b_i$ | $e_i$ | F1-score (Job Family) | F1-score (Role) |
|---|---|---|---|---|---|---|---|
| 1 | • | • | • | • | • | 0.9012 | 0.7596 |
| 2 | • | • | • |   | • | 0.9010 | 0.7598 |
| 3 | • |   | • | • | • | 0.8988 | 0.7693 |
| 4 | • |   | • |   | • | 0.8985 | 0.7518 |
| 5 | • | • | • |   |   | 0.8962 | 0.7595 |
| 6 | • | • |   | • | • | 0.8943 | 0.7653 |
| 7 | • |   | • | • |   | 0.8936 | 0.7461 |
| 8 | • | • |   | • |   | 0.8932 | 0.7662 |
| 9 | • | • |   |   | • | 0.8925 | 0.7472 |
| 10 | • |   |   | • | • | 0.8919 | 0.7518 |
| 11 | • |   | • |   |   | 0.8912 | 0.7593 |
| 12 | • | • |   | • |   | 0.8908 | 0.7507 |
| 13 | • | • |   |   |   | 0.8903 | 0.7533 |
| 14 | • |   |   |   | • | 0.8890 | 0.7540 |
| 15 | • |   |   | • |   | 0.8847 | 0.7519 |
| 16 | • |   |   |   |   | 0.8802 | 0.7539 |
| 17 |   | • | • | • | • | 0.4892 | 0.4747 |
| 18 |   | • | • |   | • | 0.4712 | 0.4703 |
| 19 |   |   | • | • | • | 0.4457 | 0.4709 |
| 20 |   |   | • |   | • | 0.4121 | 0.4688 |
| 21 |   | • | • | • |   | 0.3472 | 0.4609 |
| 22 |   | • | • |   |   | 0.3425 | 0.4613 |
| 23 |   | • |   | • | • | 0.3306 | 0.4707 |
| 24 |   | • |   |   | • | 0.3115 | 0.4710 |
| 25 |   |   | • | • |   | 0.2760 | 0.4625 |
| 26 |   |   | • |   |   | 0.2740 | 0.4613 |
| 27 |   |   |   | • | • | 0.2650 | 0.4713 |
| 28 |   |   |   |   | • | 0.2229 | 0.4713 |
| 29 |   | • |   | • |   | 0.2185 | 0.4671 |
| 30 |   | • |   |   |   | 0.1955 | 0.4625 |
| 31 |   |   |   | • |   | 0.1538 | 0.4630 |

4.1.3. Alignment of semantic representations with existing Job Family structure

Building upon the established alignment with formal HR categories, we conducted an in-depth analysis to further validate how well the semantic embedding space reflects the existing job family structure. As visualized in Figure 3(a), K-Means clustering (K=5) performed on the embeddings—without any label information—yielded clusters that show a remarkable correspondence to the actual job family distributions.

This visual observation is quantitatively supported by the similarity matrix in Figure 3(b), which shows that intra-Job Family similarities (diagonal values, all > 0.80) are consistently and significantly higher than inter-Job Family similarities. Furthermore, the matrix captures logical functional adjacencies, such as the high similarity (0.79) between the 'research' and 'clinical development' Job Families. These findings demonstrate that the proposed semantic embedding is a powerful feature capable of reconstructing the organization's formal functional structure with high fidelity, purely from the data and without explicit labels.

The exploratory analysis in Section 5.1 demonstrates that the proposed bottom-up features have proven their validity by successfully reproducing the existing static organizational concepts (Job Families). Moreover, the analysis goes beyond simple replication to reveal a dimensional network of relationships that was difficult to grasp with static data alone. In conclusion, this shows that the features developed in this study can function as powerful inputs for an advanced GNN, as they maintain alignment with existing HR information (ensuring validity) while simultaneously preventing the oversight of potential candidates not visible through formal channels.

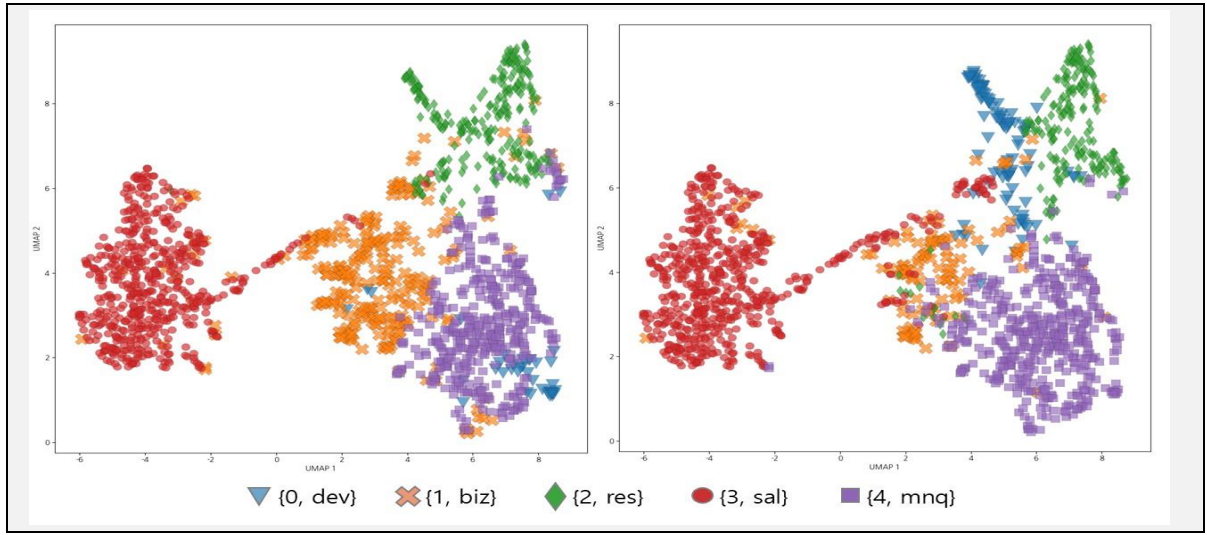

Figure 3. (a) UMAP of semantic embeddings : {k-means cluster labels,  actual Job Family labels[2]}

|     | res  | biz  | dev  | sal  | mnq  |
| --- | ---- | ---- | ---- | ---- | ---- |
| res | 0.84 | 0.77 | 0.79 | 0.73 | 0.77 |
| biz | 0.77 | 0.81 | 0.76 | 0.78 | 0.75 |
| dev | 0.79 | 0.76 | 0.82 | 0.75 | 0.74 |
| sal | 0.73 | 0.78 | 0.75 | 0.85 | 0.70 |
| mnq | 0.77 | 0.75 | 0.74 | 0.70 | 0.80 |

Figure 3. (b) Cross-Job Family semantic similarity matrix

The exploratory analysis in Section 4.1 demonstrates that the proposed bottom-up features have proven their validity by successfully reproducing the existing static organizational concepts (Job Families). Moreover, the analysis goes beyond simple replication to reveal a dimensional network of relationships that was difficult to grasp with static data alone. In conclusion, this shows that the features developed in this study can function as powerful inputs for an advanced GNN, as they maintain alignment with existing HR information (ensuring validity) while simultaneously preventing the oversight of potential candidates not visible through formal channels.

**4.2. Performance comparison of GNN-based models**
Based on the preceding analysis, this study experimented with GNN-based recommendation models that jointly utilize both semantic and structural features. Performance was evaluated using the Hit@K (K=30, 100) metric, which measures the appropriateness of talent recommendations in a rank-based manner.

Table 9 compares the talent recommendation performance (Hit@K) of the score-based baseline model against six GNN-based fusion models. The experimental results reveal a clear "staircase of success," a pattern where performance systematically improves with the increasing complexity and sophistication of the models.

Ultimately, the Late Fusion-Gating model demonstrated the most superior performance, achieving a Hit@30 of 16.1% and a Hit@100 of 40.9%, outperforming all comparative models. Notably, even the simplest single GCN model performed more than three times better than the heuristic baseline (Hit@100 ≈ 7.4%). This suggests that the GNN architecture effectively learns complex network relationships and non-linear interactions that linear models fail to capture.

---

[2] dev: clinical development, biz: business administration, res: research, sal: sales and marketing, mnq: manufacturing and quality.

Table 9. Comparison of performance

| Fusion strategy | Hit@30 | Hit@100 |
|---|---|---|
| Late Fusion (Gating) | 16.1% | 40.9% |
| Late Fusion (Attention) | 13.4% | 34.5% |
| Late Fusion (Weighted Sum, 8:2) | 12.8% | 33.5% |
| Late Fusion (Concat) | 13.4% | 34.5% |
| Early Fusion (Concat) | 10.5% | 31.4% |
| Single GCN (Plain or Semantic) | 8.1% | 27.4% |
| Heuristic Baseline 8:2 | 2.3% | 7.4% |

Furthermore, the simple fusion models that physically combined the two information sources (Early Fusion, Late Fusion-Concat) consistently outperformed the single GCN, confirming the distinct advantage of using both information sources in tandem. Taking this a step further, the adaptive fusion models (Gating), which learn the fusion methodology from the data rather than relying on fixed rules, surpassed all other models. The fact that the gating mechanism outperformed attention, in particular, suggests that for this problem, precisely controlling the information flow for each feature channel (feature-level) is a more effective strategy than adjusting the overall importance of the information sources (node-level).

### 4.3. Discussion

The experimental results empirically demonstrate that a GNN architecture that first learns semantic (the WHAT) and structural (the HOW) relationships separately and then adaptively fuses them is the most effective approach for the internal talent recommendation problem. While the score-based heuristic and single GNN models failed to fully leverage the complementary nature of these two information sources, the Late Fusion-Gating model effectively integrated them to achieve the highest recommendation performance.

The success of the top-performing gating fusion model extends beyond mere numerical superiority; it offers significant insights into model interpretability. The gating model's outstanding performance is attributed to its ability to learn distinct fusion strategies tailored to the data characteristics of each job function. An in-depth analysis of the learned gate (g) values revealed that the model undertakes a highly rational decision-making process, assigning weights to the most discriminative information within each job family.

For example, in the 'sales and marketing' function, semantic (WHAT) information exhibited high homogeneity and thus low discriminative power, as confirmed in the analysis in Section 5.1. In contrast, the structural (HOW) information was characterized by "sparsity", with most nodes being isolated while a few possessed high connectivity. The gating model identified this, assigning an overwhelming 88% weight to the sparse yet decisive structural information, rather than the less discriminative semantic information. Conversely, for the 'research' function, the model recognized the dual importance of unique expertise (WHAT) and peer collaboration (HOW), adopting a balanced strategy by utilizing the two information sources in an approximately 56:44 ratio.

This result aligns with typical organizational practices, where research positions place relatively greater emphasis on specialized academic or technical qualifications, often requiring advanced degrees, while sales and marketing job families may rely more heavily on relational networks and influence. Through this, the model demonstrates its ability to learn nuanced, context-specific fusion strategies that mirror real world talent dynamics.

Furthermore, for the 'leader' role, the model did not adhere to a fixed strategy, showing a large variance in gate values. This indicates that the model is dynamically searching for the optimal fusion method based on individual contexts, such as the leader's specific departmental affiliation. As such, the gating model presents the potential to be elevated into a recommendation system that also functions as an interpretable tool offering insights into the hidden operational principles of an organization.

Polished version of master's thesis : *Multi-View Graph Convolution Network for Internal Talent Recommendation Based on Enterprise Emails;* as of 28 July 2025
Soo Hyun Kim, Department of Applied Data Science, The Graduate School, Sungkyunkwan University; Supervised by Jang Hyun Kim (Major Advisor)
A Master' s Thesis Submitted to the Department of Applied Data Science and the Graduate School of Sungkyunkwan University in partial fulfilment of the requirements for the degree of Master of Engineering in Applied Data Science, April 2025### 4.4. Limitations and future work

This study has several limitations. While a Hit@100 performance of approximately 41% demonstrates practical utility, it also underscores the inherent complexity of the internal talent recommendation task. The limitations of the proxy labels used in our experiments and the existence of non-quantifiable factors such as reputation and growth potential need to be addressed in future research.

Therefore, we believe the following technical extensions would be valuable for learning more sophisticated organizational representations in the future:

First is the introduction of Role-aware Structural Embedding. By employing techniques like Struc2Vec, the model could learn structural role patterns—such as a 'broker' or a 'hub'—that a node performs within the network, rather than relying on simple connectivity. This would enable a more precise capture of functional similarity. For instance, a suitable successor for a departing employee who frequently acted as a 'broker' among teams would likely be another individual who performs the same 'broker' role.

Second is the extension of the model to consider the type and context of relationships. Relationships within an organization are diverse, including 'intra-team collaboration' and 'inter-departmental collaboration.' Applying an architecture like a Relational GCN (R-GCN) would allow the model to learn different information propagation methods according to the characteristics of each relationship type, thereby incorporating richer context. For example, when searching for a successor whose role was critically dependent on close collaboration with a specific team, a higher weight could be assigned to connections with that particular team.

Third is the application of time-series analysis to reflect the dynamic nature of organizations. By learning from networks that change over time using Dynamic GNNs, more timely recommendations become possible, for instance, by giving greater weight to the most recent interactions. Interaction patterns just before an employee's departure could be a more crucial clue than records from several years prior.

These extensions provide a foundation for evolving this research into a high-dimensional recommendation system that comprehensively considers organizational roles, influence, and temporal dynamics.

## Chapter 5. CONCLUSION

This study sought to redefine the problem of internal talent recommendation, moving beyond a reliance on static job titles and past histories to a data-driven approach that reflects actual interactions and semantic work content. To this end, we proposed a recommendation model that constructs both structural and semantic networks from email data, learns from each graph view independently, and then integrates them through various GNN-based fusion strategies.

Experimental results demonstrated that a model combining a dual-graph structure with a Gating fusion strategy achieved the highest recommendation performance. This empirically shows the importance of the complementary fusion of semantic and structural information for accurately judging position fit in a real-world organizational context. Notably, considering the nature of the HR environment where explicit labels are absent, this study designed a weak supervision framework and provided explainable recommendations based on diverse factors, including semantic similarity and structural centrality.

This thesis presented a framework capable of inferring suitable talents based on similarity in a label-free environment. Furthermore, it confirmed the potential for applying various GNN extension techniques—such as structural role embeddings, influence-based information propagation, and time-series network analysis—for more sophisticated organizational representation learning in the future. This indicates the potential for this work to evolve beyond talent recommendation into an intelligent system applicable to the broader spectrum of decision-making within the HR domain.

**초록(국문)**

**내부 인재 추천을 위한 이메일 네트워크 기반 다중 관점 그래프 합성곱 신경망**

내부 인재 추천은 조직의 연속성을 위한 핵심 전략이지만, 후보자 풀링 시 소수 관리자의 시야에 의존하여 잠재적 포지션 적합 후보자가 누락되는 구조적 한계를 가진다. 본 연구는 이러한 문제를 해결하기 위해, 조직 내 이메일 데이터로부터 WHAT(업무의 의미적 유사성)과 HOW (상호작용의 구조적 관계)를 각각 독립적인 그래프로 모델링하고, 이를 게이팅 메커니즘 (gating mechanism)을 갖춘 이중 GCN을 통해 적응적으로 융합하는 새로운 프레임워크를 제안한다. 실험 결과, 제안된 게이팅 기반 융합 모델은 다른 융합 전략들과 휴리스틱 Baseline을 모두 능가하며 Hit@100 기준 40.9%의 가장 우수한 성능을 달성했다. 특히, 학습된 게이트 값은 '영업' 직군처럼 관계적 (HOW) 정보가 중요한 상대적으로 중요한 직군과 덜 중요한 직군을 구분하여, 각기 다른 융합 전략을 스스로 학습하는 높은 수준의 해석 가능성 (interpretability)을 보여주었다. 이 연구는 전통적인 방법에 내재된 후보자 누락의 위험을 완화하면서, 내부 인재 탐색을 위한 정량적이고 포괄적인 프레임워크를 제공한다. 또한, 본 논문은 서로 다른 조직 포지션에서의 성공을 위해 요구되는 업무 적합성 (WHAT)과 협업 패턴 (HOW) 간의 최적 결합 비율을 경험적으로 규명할 수 있도록 하는 실무적 및 학문적 기여점이 있다.

**키워드: 조직 네트워크 분석, 그래프 딥러닝, 인재 추천, 후임자 추천, 피플 애널리틱스**